\documentclass[10pt,twocolumn,letterpaper]{article}

\usepackage{cvpr}              
\definecolor{cvprblue}{rgb}{0.21,0.49,0.74}
\usepackage[pagebackref,breaklinks,colorlinks,allcolors=cvprblue]{hyperref}
\usepackage{algorithm}
\usepackage{algorithmic}

\def\paperID{30096} 
\def\confName{CVPR}
\def\confYear{2026}

\newcommand{\methodname}{\textit{FocaLogic}}

\title{\methodname: Logic-Based Interpretation of Visual Model Decisions}

\author{Chenchen Zhao* \quad Muxi Chen* \quad 
Qiang Xu$^\dagger$ \\
Department of Computer Science and Engineering, The Chinese University of Hong Kong \\
{\tt\small cczhao@cse.cuhk.edu.hk \quad mxchen21@cse.cuhk.edu.hk \quad qxu@cse.cuhk.edu.hk} \\
{\small * Equal contribution \quad $^\dagger$ Corresponding author}
}
\begin{document}

\maketitle

\begin{abstract}
Interpretability of modern visual models is crucial, particularly in high-stakes applications. However, existing interpretability methods typically suffer from either reliance on white-box model access or insufficient quantitative rigor. To address these limitations, we introduce \methodname, a novel model-agnostic framework designed to interpret and quantify visual model decision-making through logic-based representations. \methodname~identifies minimal interpretable subsets of visual regions—termed \emph{visual focuses}—that decisively influence model predictions. It translates these visual focuses into precise and compact logical expressions, enabling transparent and structured interpretations. Additionally, we propose a suite of quantitative metrics, including focus precision, recall, and divergence, to objectively evaluate model behavior across diverse scenarios. Empirical analyses demonstrate \methodname's capability to uncover critical insights such as training-induced concentration, increasing focus accuracy through generalization, and anomalous focuses under biases and adversarial attacks. Overall, \methodname~provides a systematic, scalable, and quantitative solution for interpreting visual models.
\end{abstract}

\begin{figure*}[t]
\centering
\includegraphics[width=0.96\textwidth]{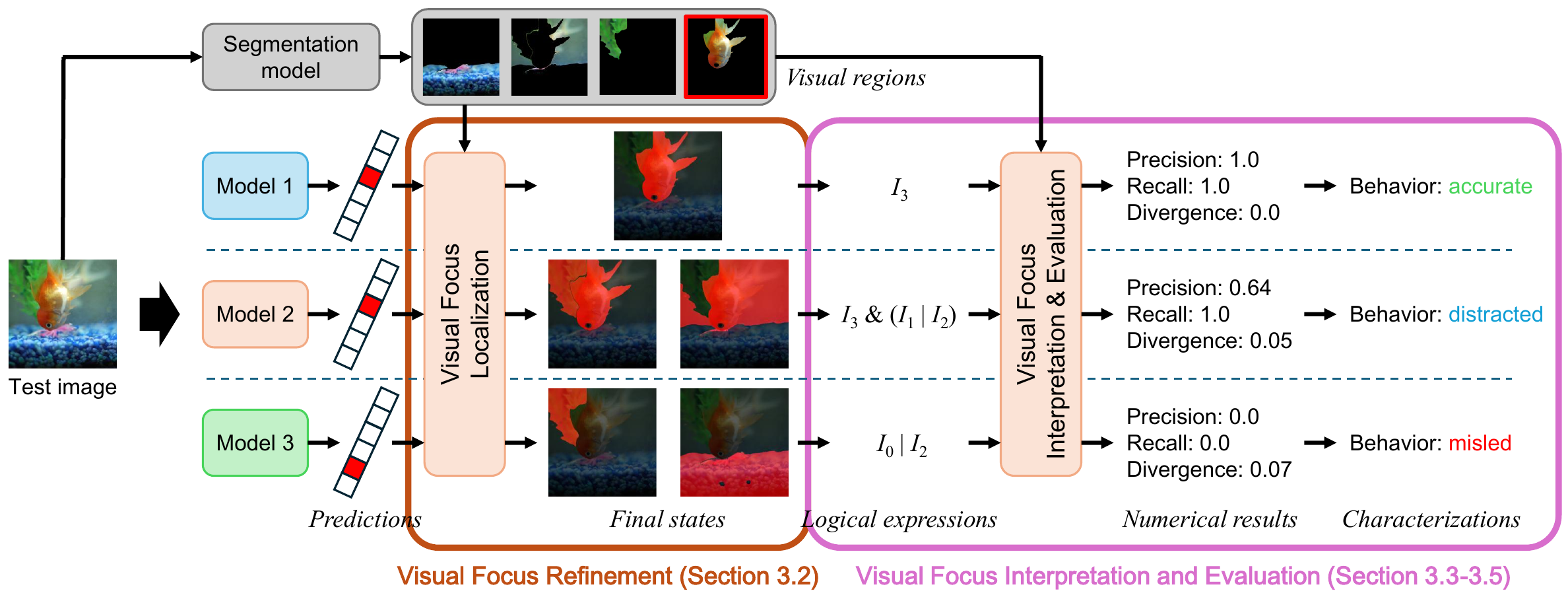}
\caption{The overall workflow of \methodname. Given a target model and an input image, \methodname~first localizes key visual regions that constitute the visual focus of the model on the image. It then derives a logical expression from the visual focus to interpret the model's decision-making behavior. Finally, it introduces a comprehensive set of metrics to evaluate the model's performance on the image from a focus-based perspective. $I_i$ denotes the $i^\text{th}$ visual region of image $I$.}
\vspace{-0.2cm}
\label{fig:teaser}
\end{figure*}

\section{Introduction}\label{sec:intro}

Model interpretability has emerged as a critical concern in modern machine learning~\cite{zhang2018visual,bereska2024mechanistic}, particularly with the growing deployment of deep neural networks in high-stakes domains such as healthcare~\cite{buolamwini2018gender} and autonomous systems~\cite{fujiyoshi2019deep}. A central aspect of interpretability is understanding a model's decision focus—identifying the most influential parts of an input that drive model predictions~\cite{selvaraju2016grad}. Such insights are essential for ensuring mechanistic transparency, diagnosing erroneous behavior, and uncovering potential biases in real-world applications.

Numerous interpretability techniques~\cite{chen2023hibug,cunningham2023sparse,koh2020concept} have been proposed, yet many rely on white-box access to the model and require internal information such as gradients~\cite{selvaraju2016grad,selvaraju2017grad} or activation maps~\cite{chen2020adapting}. However, state-of-the-art (SOTA) models—especially those used in commercial settings~\cite{hurst2024gpt}—are often proprietary, highly complex, and closed-source, making white-box interpretability methods infeasible in practice. As an alternative, model-agnostic (i.e., black-box) approaches~\cite{chen2025hibug2,casalicchio2018visualizing,covert2021explaining} provide interpretability without requiring access to internal model architectures. Yet, these methods typically rely on auxiliary surrogate models~\cite{ribeiro2016should}, introducing computational overhead and extra opacity. Moreover, their outputs often lack formal structures, depend heavily on subsequent human inspection, and struggle to scale due to the absence of standardized, quantitative evaluation metrics.

To address these limitations, we propose \methodname, a novel model-agnostic approach for localizing, interpreting, and evaluating key visual regions that drive model predictions—referred to as \textit{visual focuses}. Given an input image segmented into multiple visual regions, \methodname~iteratively refines these regions to identify minimal subsets that preserve the model's original prediction. These subsets are then translated into a precise logical expression, which compactly encodes the model's decision-making criteria. Then, we introduce a set of metrics—including \textit{precision}, \textit{recall} and \textit{divergence}—to quantitatively evaluate model focuses.

We conduct extensive comparisons between \methodname~and other visual focus interpretation methods to validate its superiority in \textbf{decisive and comprehensive interpretation}, \textbf{sensitivity to behavior anomalies}, and \textbf{semantic alignment}. Furthermore, using \methodname, we uncover several empirical insights into model behavior, including but not limited to: (a) models gradually concentrate during training—as training progresses, models tend to develop more concentrated focuses, increasingly relying on semantically meaningful visual regions; (b) models focus more precisely through generalization—models trained on broader, larger-scale tasks exhibit more precise focuses than those trained on narrow sub-tasks, even when achieving comparable test accuracies; (c) models are easily distracted under biases and adversarial attacks—biased or adversarial inputs often lead models to attend to irrelevant or spurious regions, exposing hidden dependencies not captured by standard evaluation metrics.

The main contributions of this paper are:
\begin{enumerate}
\item \textbf{Logic-based model focus interpretation}: we propose \methodname, a model-agnostic method that translates visual focuses into precise logical expressions, enabling structured representations of model decisions.
\item \textbf{Quantitative metrics for focus analysis}: we introduce a comprehensive set of metrics, including \textit{precision}, \textit{recall}, and \textit{divergence}, that facilitate automated, large-scale evaluation of model focuses.
\item \textbf{Empirical insights into model behavior}: through extensive experiments, we demonstrate that: (a) model focuses become more concentrated as training progresses; (b) better generalization leads to more accurate focuses; and (c) dispersed or inconsistent focuses often correlate with biased or adversarial inputs, which other methods of the same kind fail to reveal.
\end{enumerate}


\begin{table*}[t]
\centering
\small
\caption{Characteristics of the related methods and \methodname.}
\begin{tabular}{c|c|c|c|c|c|c|c}
\toprule
Characteristics & SHAP & FR & GradCAM & LIME & VisionLogic & AND-OR & \methodname~(ours)\\
\midrule
Visual focus interpretation & & & $\checkmark$ & $\checkmark$ & $\checkmark$ & $\checkmark$ & $\checkmark$\\
Model-agnostic & $\checkmark$ & $\checkmark$ & & $\checkmark$ & & $\checkmark$ & $\checkmark$\\
Training-free & $\checkmark$ & $\checkmark$ & $\checkmark$ & & $\checkmark$ & & $\checkmark$\\
Scalable automated analysis & & & & & & & $\checkmark$\\
Numerical interpretation & & & & & & & $\checkmark$\\
\bottomrule
\end{tabular}
\vspace{-0.2cm}
\label{tab:related}
\end{table*}

\section{Related Work}\label{sec:related_work}

\subsection{Behavioral Interpretations of Neural Networks}
Recent work~\cite{bereska2024mechanistic} categorizes model interpretability methods into four types: behavioral~\cite{selvaraju2017grad}, attributional~\cite{shrikumar2017learning}, concept-based~\cite{koh2020concept}, and mechanistic~\cite{dunefsky2024transcoders}. Among these, behavioral interpretation is distinct in its support for black-box analysis, as it does not require access to models' internal components. \methodname~falls within the behavioral category.

Behavioral methods generally analyze input-output relationships by perturbing or pruning input segments and observing changes in model predictions. For example, SHAP~\cite{casalicchio2018visualizing} perturbs local and global features and proposes Shapley Feature IMPortance (SFIMP) to estimate each feature's contribution to model output. Feature Removal (FR)~\cite{covert2021explaining} removes specific features and assesses resulting prediction shifts.
However, these approaches largely focus on individual features or isolated perturbations, overlooking the combinatorial interactions among correlated features, which are often critical for understanding complex model behavior. Additionally, these methods lack intuitive mechanisms for representing models' focus patterns, particularly in the visual domain.

\subsection{Analysis of Visual Model Focuses}

Grad-CAM is the most widely adopted approach for interpreting visual model attention. However, its reliance on features and gradients makes it inapplicable to model-agnostic scenarios. LIME~\cite{ribeiro2016should}, a model-agnostic alternative, approximates visual focuses by training local surrogate models. While broadly applicable, LIME suffers from high computational overhead, training inaccuracies, and limitations in generalizing across diverse samples or tasks. VisionLogic~\cite{geng2025learning} adopts a white-box approach by converting final-layer activations into logical predicates, which are then grounded to visual concepts through human annotation. However, it requires access to intermediate features and lacks comprehensive evaluations of model behavior.

More recently, AND-OR interaction-based approaches~\cite{ren2024towards,ren2025monitoring}
have emerged as promising alternatives. They explicitly model interactions among input segments using learnable AND/OR operators in a model-agnostic fashion. Although offering more structured explanations, they remain learning-based and are thus susceptible to regression errors, mechanistic transparency of the learned operators, and high training complexity.

More importantly, all of the aforementioned methods share a fundamental limitation: they rely heavily on human inspection of individual explanations, making it difficult to infer \textbf{\textit{global}} model behavior across large-scale datasets. While some techniques, such as SHAP, produce quantitative outputs, they lack automated mechanisms for precise and comprehensive evaluation.

To address these limitations, \methodname~is designed as a training-free and fully-automated framework that enables scalable behavioral analysis with systematic numerical evaluation. Table~\ref{tab:related} summarizes the key characteristics of existing methods in comparison to \methodname.
\section{Localizing, Interpreting, and Evaluating Visual Focuses of Neural Networks}\label{sec:method_localize}

\subsection{Overview}\label{sec:method_localize_overview}

Figure~\ref{fig:teaser} illustrates the overall workflow of \methodname. The process consists of four key stages: visual focus refinement, logic-based focus translation, numerical focus evaluation, and focus-driven model behavior characterization. 

\begin{figure}[t]
\centering
\includegraphics[width=0.96\linewidth]{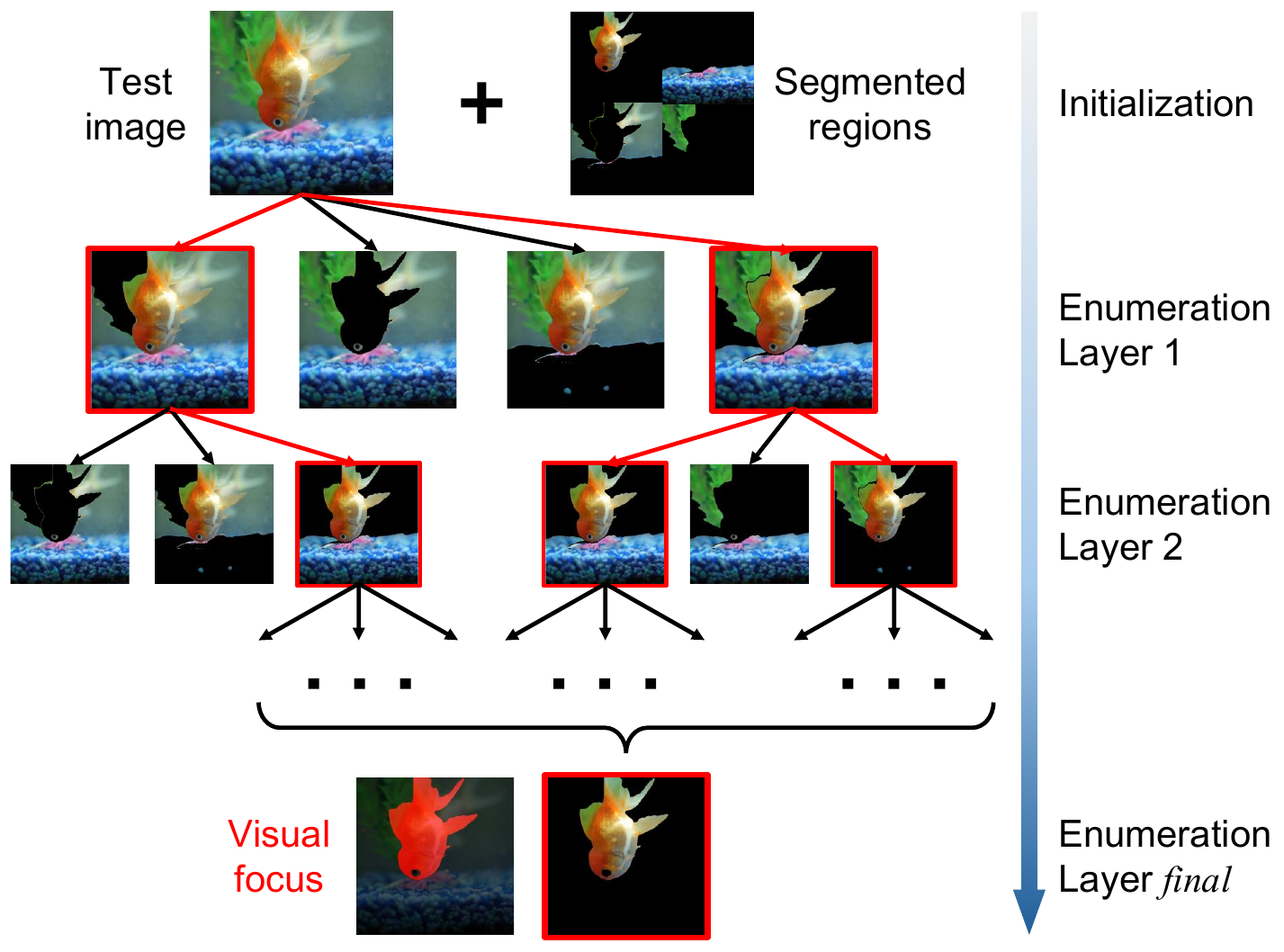}
\caption{An example of visual focus refinement.}
\vspace{-0.2cm}
\label{fig:enum}
\end{figure}

\subsection{Enumerative Visual Focus Refinement}\label{sec:method_enum}

Given an image, \methodname~first performs global segmentation (using SAM~\cite{kirillov2023segment} in this paper) to obtain multiple visual regions corresponding to different parts of the objects and the background. Then, given a target model, it iteratively prunes irrelevant regions to localize the most influential ones, which are considered the visual focuses of the model for the image.

For an image $I$ with $M$ visual regions, \methodname~initializes a state vector $v^\text{init}=\mathbf{1}^M$, indicating that all regions are initially preserved. Then, given a target model $f$, \methodname~refines the visual focus through the following iterative steps, applied to each current state $v$:
\begin{enumerate}
\item Region pruning: let $v$ have $n$ preserved regions with indices $i_1,i_2,\cdots,i_n$. Derive $n$ new candidate states $v^1,v^2,\cdots,v^n$ from $v$ by individually removing one preserved region from $v$: $v^1_{i_1},v^2_{i_2},\cdots,v^n_{i_n}\leftarrow0$.
\item State validation: let $I[v]$ denote the image $I$ with specific regions pruned according to $v$. $v$ is considered valid if the model prediction remains unchanged: $f(I[v])=f(I)$. Run the target model on all $n$ states generated in region pruning, and preserve all valid ones.
\end{enumerate}
These steps are repeated for multiple rounds until no new valid state is derived, as illustrated in Figure~\ref{fig:enum}. The resulting set of final states $V$ satisfies the following properties:
\begin{itemize}
\item \textbf{Validity}: all final states preserve the model prediction: $\forall v\in V,f(I[v])=f(I)$.
\item \textbf{Minimality}: no smaller region subset remains valid: $\forall v^a\notin V,v^b\in V,||v^a||_0<||v^b||_0\Rightarrow f(I[v^a])\ne f(I)$.
\end{itemize}

In each final state, all retained regions are indispensable to model decisions. This \textbf{\textit{decisive}} property enables \methodname~to provide more rigorous behavioral interpretations than feature-guided visual explanation methods such as Grad-CAM.

\subsection{Final States to Logical Expressions}\label{sec:method_logic}


\methodname~initializes visual focus evaluation by translating the extracted visual focuses into structured logical expressions. Specifically, it recursively merges shared region indices among the final states using \textit{AND} operations, and clusters unique region indices using \textit{OR} operations. For instance, consider the following set of final states:
\begin{equation}
V=\{\{I_1,I_2,I_3,I_5\},\{I_1,I_2,I_3,I_6\},\{I_1,I_4\}\}
\end{equation}
\methodname~translates it into the logical expression:
\begin{equation}
T(V)=I_1\ \&\ (I_4\ |\ (I_2\ \&\ I_3\ \&\ (I_5\ |\ I_6)))
\end{equation}
which compactly represents the model's decision for $I$. In addition, the logic expressions also enable quantitative metrics, such as Divergence (stated in Section~\ref{sec:method_metrics}), for large-scale comprehensive analysis. The pseudo-code for the translation is provided in Appendix 1.

\subsection{Metrics for Visual Focus Evaluation}\label{sec:method_metrics}

Before evaluation, we perform label-guided segmentation to determine the ground-truth state $\bar{v}$ for each image, serving as a reference for correctness. Specifically, we utilize Grounded-SAM~\cite{ren2024grounded} with label texts as guidance, with settings consistent with SAM in visual focus refinement. We do not directly adopt text-guided segmentations as ground-truths. Instead, we match global segmentations with them using intersection-over-union (IoU) with a preset threshold (0.7 in this paper). Under this setting, the ground-truth regions are subsets of the initial visual regions, ensuring high consistency and solving possible discrepancies between raw results of global and text-guided segmentations. Note that due to the identical backbones of the two segmentation models, the empirical IoU scores are higher than 0.95 for most matched cases. Therefore, the matching threshold has minimal impact on the results.

Given the final state set $V$ and the corresponding ground-truth visual focus $\bar{v}$ for image $I$, we conduct comprehensive quantitative focus assessments using the following metrics:
\begin{itemize}
\item \textbf{Precision}, which measures whether the model focuses on the correct regions of the image:
\begin{equation}
\mathcal{P}=\frac{1}{|V|}\sum\limits_{v\in V}\frac{\mathcal{S}(I[v\cap\bar{v}])}{\mathcal{S}(I[v])}
\end{equation}
where $\mathcal{S}$ denotes the area determination function. $(I[v])$ refers to the visual focus in $I$ according to state $v$.
\item \textbf{Recall}, which measures whether the model includes all necessary regions of the image required for correct decision-making:
\begin{equation}
\mathcal{R}=\frac{1}{|V|}\sum\limits_{v\in V}\frac{\mathcal{S}(I[v\cap\bar{v}])}{\mathcal{S}(I[\bar{v}])}
\end{equation}
\item \textbf{Divergence}, which measures whether the model consistently focuses on specific regions of the image across different final states:
\begin{gather}
\mathcal{D}=||\mathcal{S}(I_{1:M})\cdot Var(V)||_1\\
Var(V)=\{Var_i(V)|i\in\{1,2,\cdots,M\}\}\\
Var_i(V)=\sigma(\{v_i|v\in V\})
\end{gather}
where $\sigma$ denotes the variance function, $\mathcal{S}(I_{1:M})$ is the area vector of all $M$ regions. $\mathcal{D}$ reflects the area-weighted variance of region selections across all final states.
\end{itemize}

Note that high divergence does not necessarily imply poor performance. For instance, in image containing multiple instances of the same class, a model may focus on any of them and still produce a correct prediction. Therefore, divergence should be interpreted in conjunction with precision and recall for a holistic assessment of model behavior.

\begin{figure}[t]
\centering
\includegraphics[width=0.8\linewidth]{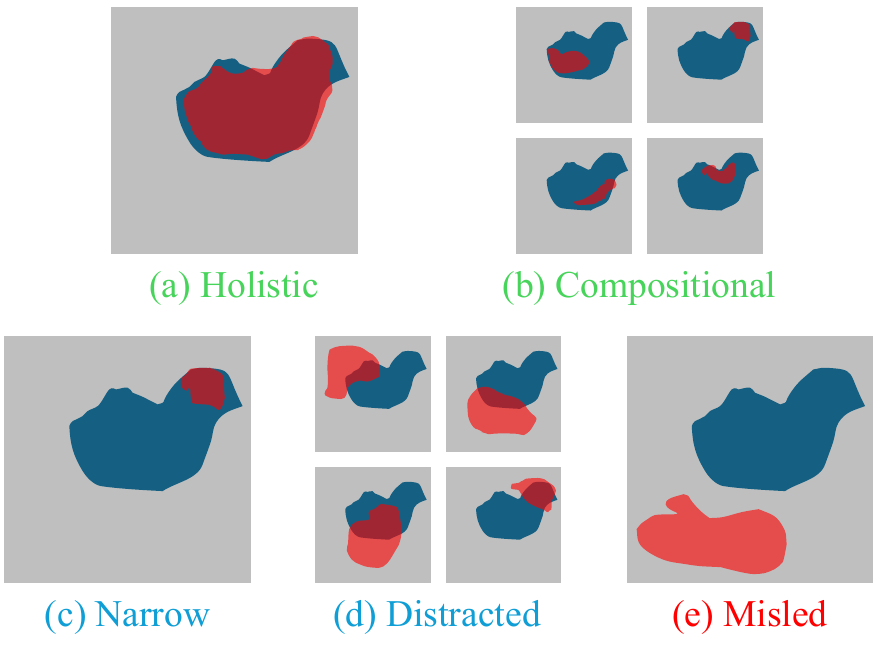}
\caption{Common types of visual model behavior from the focus-based perspective, with the ground-truth focus in blue and the model's visual focus in red.}
\vspace{-0.2cm}
\label{fig:interpretation}
\end{figure}

\subsection{Focus-Based Behavior Interpretation}\label{sec:method_interpretations}


From a focus-based perspective, we categorize typical visual model behavior into five distinct types, as illustrated in Figure~\ref{fig:interpretation}. Based on the observed behavior patterns, we propose the following metric-driven criteria to quantitatively characterize global model behavior:

\begin{itemize}
\item \textbf{Holistic} = high precision + high recall + low divergence. The model focuses on the correct regions holistically.
\item \textbf{Compositional} = high precision + low recall + high divergence. The model captures multiple correct sub-regions for compositional recognitions.
\item \textbf{Narrow} = high precision + low recall + low divergence. The model attends to only a small portion of the correct regions. While possibly accurate, such behavior may suffer from stability issues in complex scenarios.
\item \textbf{Distracted} = low precision + high recall. The model includes the correct regions but also attends to multiple irrelevant areas, typically with high divergence.
\item \textbf{Misled} = low precision + low recall. The model focuses entirely on incorrect regions, potentially indicating severe behavioral anomalies.
\end{itemize}
Note that an accurate visual focus does not necessarily indicate a correct model prediction, and vice versa.

To our knowledge, \methodname~is among the first works to provide fine-grained, numerical, and systematic analysis of visual focuses and their implications for model behavior.

\subsection{Techniques for Efficiency}\label{sec:method_efficiency}

State enumeration in visual focus refinement poses significant computational challenges due to exponential time complexity. To address this, we propose several techniques to improve the efficiency of the enumeration process.
\begin{itemize}
\item Segmentation simplification: before enumeration, we perform segmentation with larger receptive fields (by reducing \textit{points\_per\_side} in SAM). This reduces the average number of visual regions by 20\%-40\%, with minimal impact on segmentation quality.
\item Region merging: we set a minimum area proportion threshold of $10^{-3}$, which effectively limits the number of regions for enumeration and improve the quality of each visual region simultaneously. All regions with area proportions below the threshold, along with all unsegmented areas, are merged into a single additional region to ensure complete image coverage.
\item Beam-search-style state pruning. Inspired by the beam search~\cite{bisiani1992beam} technique widely-applied in language generation, we derive at most $k$ new states from each current valid state. As demonstrated in Section~\ref{sec:exp_scalability}, selecting an appropriate $k$ significantly reduces inference time without compromising the accuracy of \methodname.
\end{itemize}
\section{Experiments}\label{sec:exp}

\subsection{Experimental Setup}

In this section, we first compare \methodname~with two widely used visual focus interpretation baselines, GradCAM~\cite{selvaraju2017grad} and LIME~\cite{ribeiro2016should}, under multiple scenarios. Then, we use \methodname~to analyze model behavior from the perspectives of, anomalous behaviors, training dynamics, generalizability, and cross-model variations across modern architectures. Finally, we study the robustness, generality, and efficiency of \methodname.

All experiments are conducted on ImageNet. Multiple representative image classification models (ResNet18~\cite{he2016deep}, ViT-L16~\cite{dosovitskiy2020image}, and three variants of zero-shot CLIP~\cite{radford2021learning}) and multi-modal large language models (MLLM) (QWen2-VL-2B-Instruct~\cite{wang2024qwen2} and GPT-4o~\cite{hurst2024gpt}) are involved. To ensure the reliability of our interpretability analysis, beam-search-style pruning is disabled in all experiments, except in the efficiency study in Section~\ref{sec:exp_scalability}.


\subsection{Baseline Comparisons}\label{sec:exp_baseline}

\begin{figure}[t]
\centering
\includegraphics[width=0.96\linewidth]{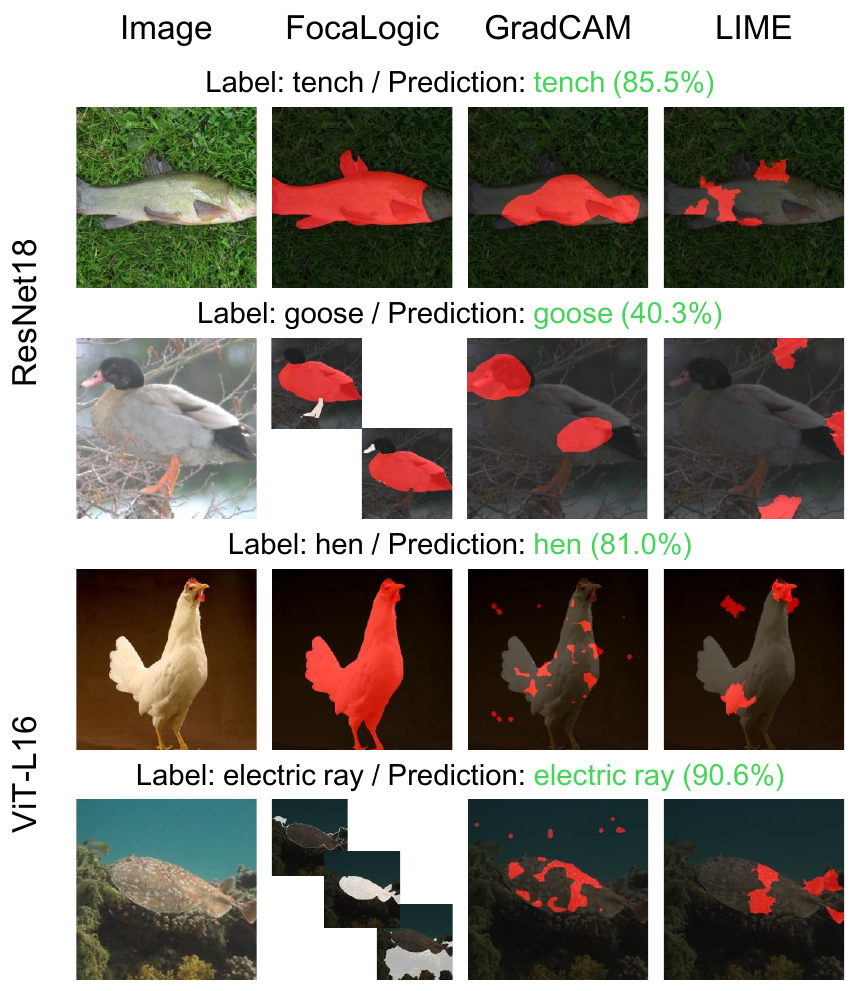}
\caption{Horizontal comparisons on the localized visual focuses of ResNet18 and ViT-L16 in common scenarios. For the baselines, the localized regions are marked red. For \methodname, regions shared among all final states are marked red, and unique regions are marked white.}
\vspace{-0.2cm}
\label{fig:horizontal_vis_baseline}
\end{figure}

With ResNet18 and ViT-L16 as examples, the horizontal comparisons between \methodname~and the baselines are presented in Figure~\ref{fig:horizontal_vis_baseline}. Results demonstrate that \methodname~surpasses baseline methods in several key aspects:
\begin{itemize}
\item \textbf{More comprehensive interpretations.} By incorporating cross-region \textit{AND} and \textit{OR} relationships, \methodname~produces multiple final states that represent complex behavior logic. In contrast, GradCAM and LIME highlight isolated regions without revealing inter-region relationships.
\item \textbf{Better semantic alignment.} With global segmentation prior to visual focus refinement, \methodname~localizes visual regions that correspond closely to semantically meaningful image components. In contrast, GradCAM suffers from low spatial precision, due to the low resolutions of deep features, while LIME's visual focuses exhibit limited spatial coherence because it performs local regressions that fail to capture sufficient global context.
\end{itemize}

\subsection{Interpretations of Anomalous Behavior}\label{sec:exp_scenarios}

In realistic scenarios, visual models are frequently exposed to biases in data, perturbations in inputs, and adversarial attacks. To rigorously evaluate how these challenging conditions influence model behaviors as captured by \methodname, we conduct the following experiments:
\begin{itemize}
\item \textbf{Inductive bias in training data}: we construct a biased training set using 10 ImageNet classes, where 3 classes (acoustic guitar, coffee mug, soccer ball) are replaced with images generated by Stable Diffusion 3.5~\cite{esser2024scaling} using biased instructions. A ResNet18 is trained on the constructed dataset to encode biased behavior. The biased model is then evaluated on unbiased test data of the same classes. Prompts for SD3.5 are detailed in Appendix 2.
\item \textbf{Noise / perturbations in test samples}: we use an ImageNet-pretrained ResNet18 and add maximum amplitudes of Gaussian noise to the test images while keeping the corresponding model predictions unchanged.
\item \textbf{Adversarial attacks on models}: Using the same pretrained ResNet18, we generate an adversarial example from each test image via targeted projected gradient descent (PGD)~\cite{madry2017towards}, with the target and the ground-truth classes in distinct super-categories. Targeted PGD achieves a 92.5\% success rate on the test set.
\end{itemize}


\begin{table}[t]
\centering
\small
\caption{Numerical comparisons of ResNet18 in different test scenarios. Results are calculated and averaged on all test images to reflect global model behavior in each scenario.}
\begin{tabular}{c|c|c|c|c}
\toprule
Scenario & Top-1 acc & $\mathcal{P}$ & $\mathcal{R}$ & $\mathcal{D}$\\
\midrule
Vanilla & 0.83 & 0.72 & 0.67 & 0.04\\
Biased & 0.70 & 0.18 & 0.25 & 0.06\\
Perturbed & 0.83 & 0.60 & 0.73 & 0.05\\
Attacked & 0.00 & 0.42 & 0.69 & 0.06\\
\bottomrule
\end{tabular}
\vspace{-0.2cm}
\label{tab:scenarios}
\end{table}

\begin{figure}[t]
\centering
\includegraphics[width=0.96\linewidth]{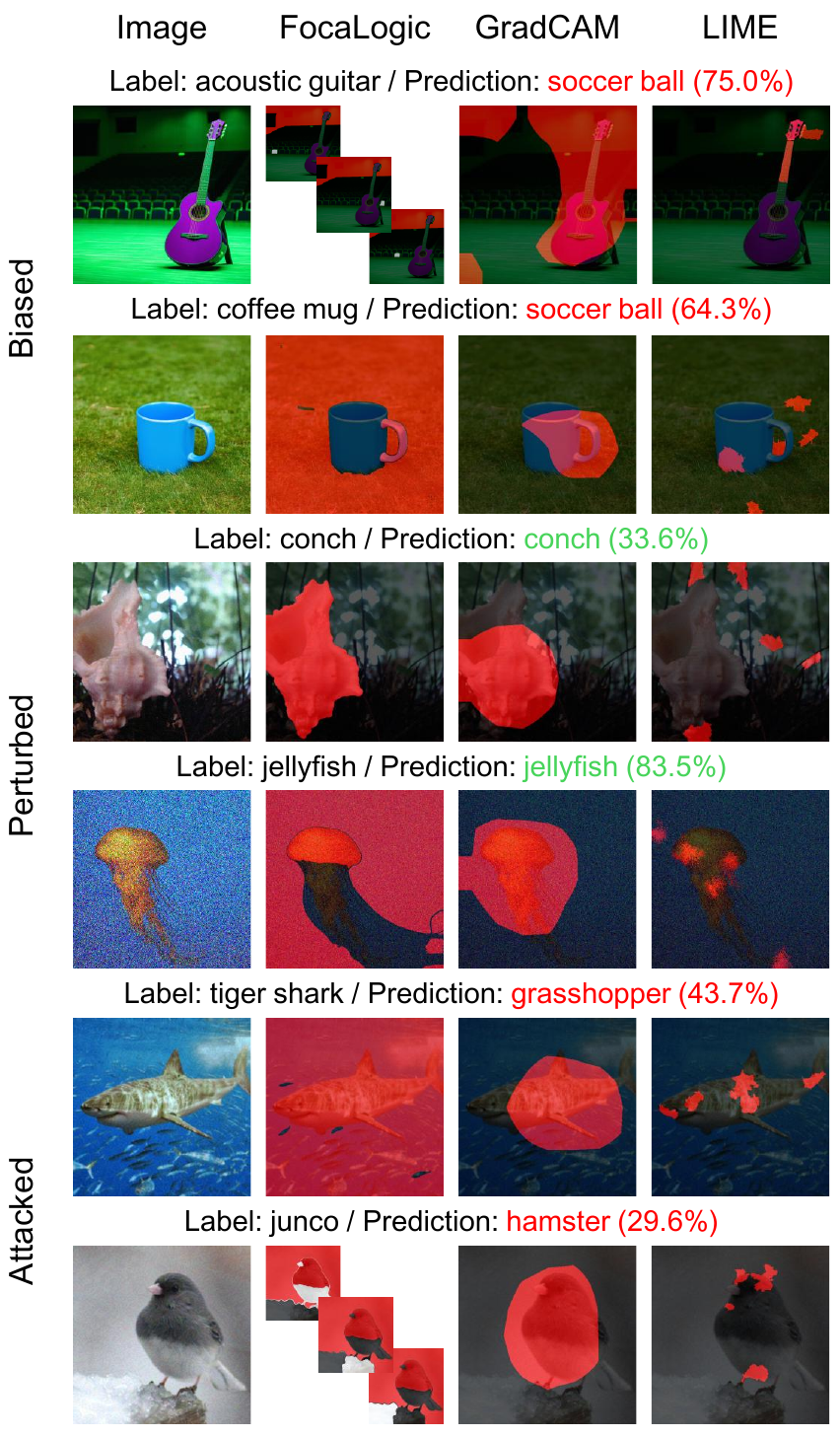}
\caption{Horizontal comparisons on the localized visual focuses of ResNet18 in different scenarios.}
\vspace{-0.2cm}
\label{fig:scenarios_vis_baseline}
\end{figure}

Table~\ref{tab:scenarios} summarizes the model behavior across different test scenarios, with comparative visualization examples shown in Figure~\ref{fig:scenarios_vis_baseline}. Results validate the superiority of \methodname~from the model error perspective:
\begin{itemize}
\item \textbf{Higher sensitivity to model errors.} \methodname~effectively reveals anomalous model behavior through its visual focuses, while GradCAM and LIME often continue to highlight the main objects even when the model exhibits biases or distraction.
\end{itemize}

Interpretations by \methodname~also reveal several anomalies in model behavior:
\begin{itemize}
\item In the biased scenario, the model shows significantly lower focus precision and recall compared to the vanilla model, indicating a shift in visual focuses towards irrelevant image content due to training biases.
\item In the perturbed scenario, the model exhibits a slight precision drop but higher recall and divergence, suggesting that noise induces distracted patterns of visual focuses. 
\item In the attacked scenario, despite a 0\% accuracy, the model maintains recall comparable to the vanilla model while showing reduced precision and higher divergence, indicating that adversarial attacks, under our setup, primarily cause distraction rather than focus misalignment. 
\end{itemize}

\subsection{Training Dynamics with Different Model Generalizability}\label{sec:generalizability}

In this experiment, we compare the behavior of two models: a specialized ResNet18 trained on 10-class ImageNet subset (same as section~\ref{sec:exp_scenarios}), and a vanilla ResNet18 trained on the full ImageNet dataset. Both models are trained from scratch for 50 epochs and evaluated on the same 10-class subset.

\begin{figure}[t]
\centering
\subfloat[Top-1 accuracy]{\includegraphics[width=0.4\linewidth]{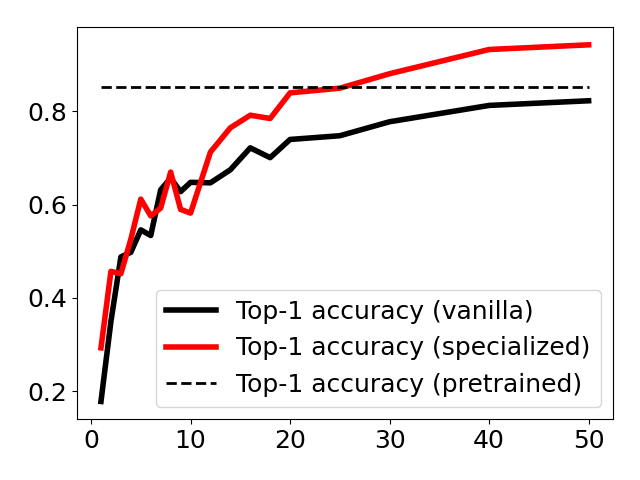}}
\subfloat[$\mathcal{P}$]{\includegraphics[width=0.4\linewidth]{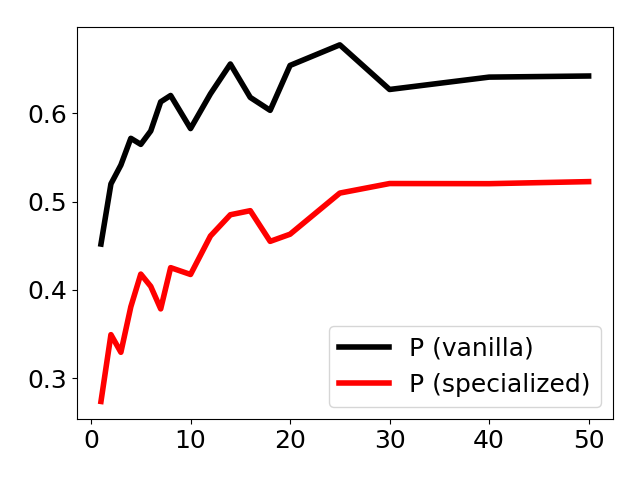}}\\
\subfloat[$\mathcal{R}$]{\includegraphics[width=0.4\linewidth]{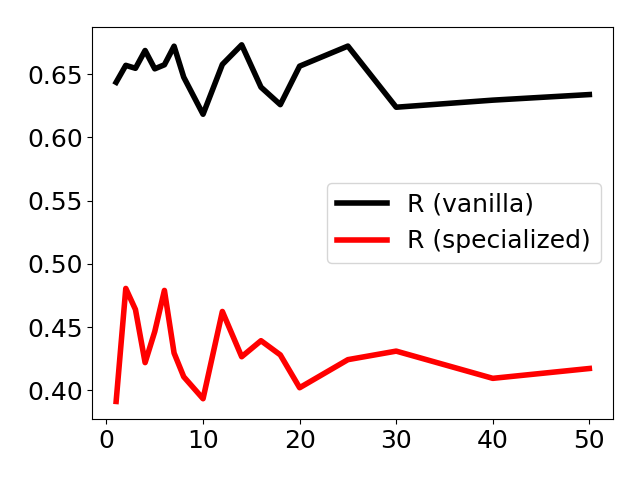}}
\subfloat[$\mathcal{D}$]{\includegraphics[width=0.4\linewidth]{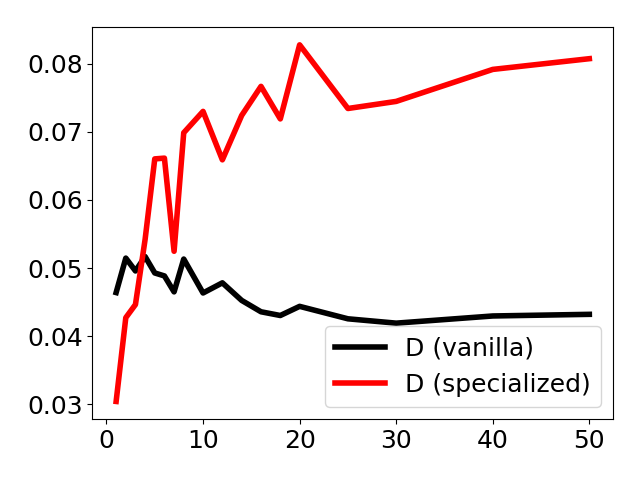}}
\caption{Metric comparisons of vanilla and specialized (trained on 10 classes) ResNet18 through 50-epoch training.}
\vspace{-0.2cm}
\label{fig:specialized}
\end{figure}

\begin{figure}[t]
\centering
\includegraphics[width=0.96\linewidth]{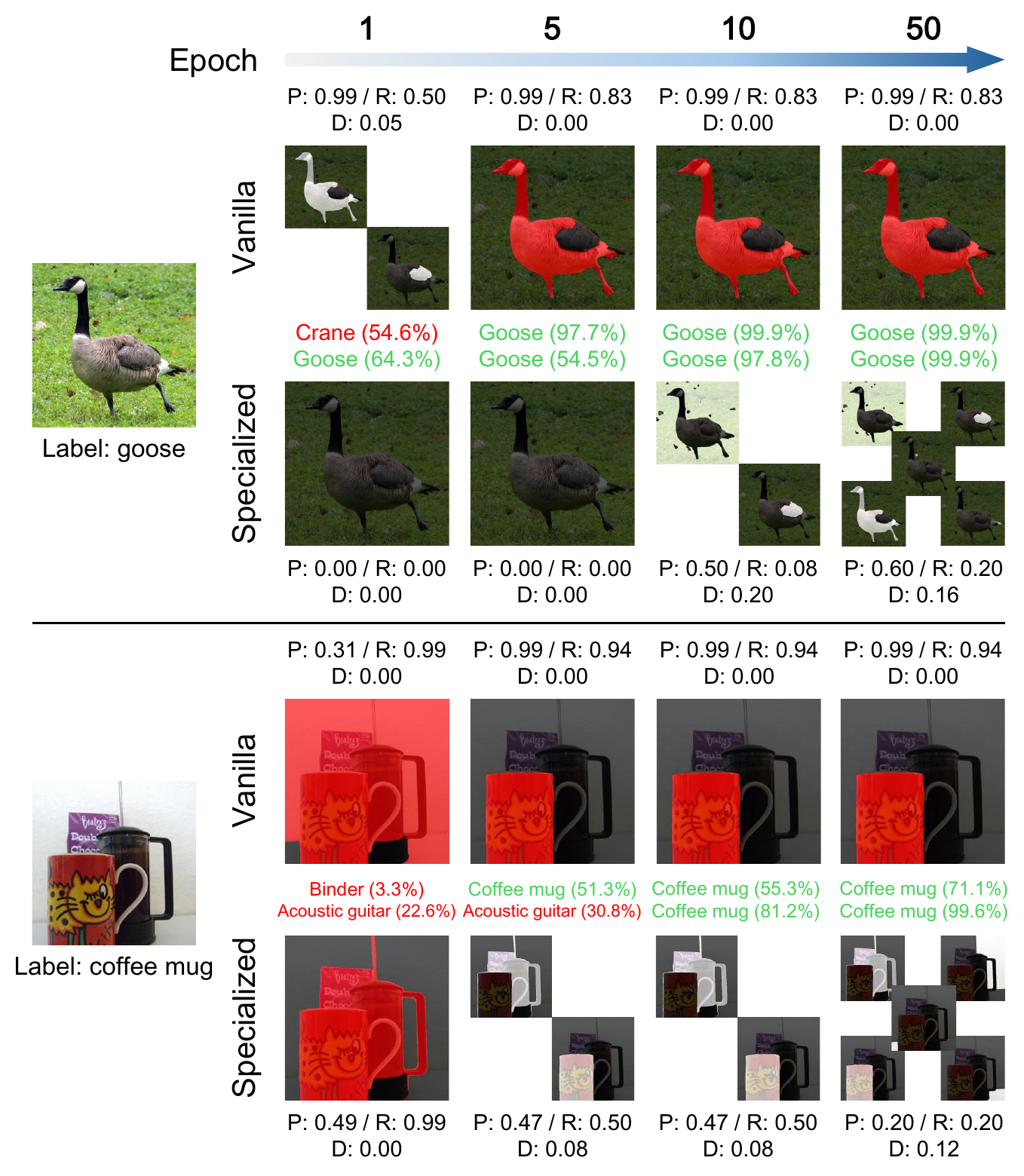}
\caption{Representative visual focuses of ResNet18 with different generalization capabilities.}
\vspace{-0.2cm}
\label{fig:specialized_vis}
\end{figure}

Figure~\ref{fig:specialized} shows the changes of model performance throughout training, with visualized qualitative comparisons between the two models shown in Figure~\ref{fig:specialized_vis}. While both exhibit similar trends in precision and recall,
the specialized model displays a steadily increasing divergence. This suggests that models with limited generalizability tend to become more distracted (i.e., focus more on irrelevant image content) as training progresses. This phenomenon can be attributed to the nature of simple classification tasks, where cross-class visual differences may arise from both the main objects and background artifacts. As a result, models trained on narrow datasets are prone to overfitting to irrelevant features that happen to correlate with class distinctions. Consequently, despite achieving higher top-1 accuracy, the specialized model exhibits lower precision and recall than its vanilla counterpart. Quantitative metric values of the model training dynamics are presented in Appendix 3.1. More visualizations of horizontal comparisons on the specialized model with relatively lower generalizability are shown in Appendix 3.2.

\begin{figure}[t]
\centering
\includegraphics[width=0.96\linewidth]{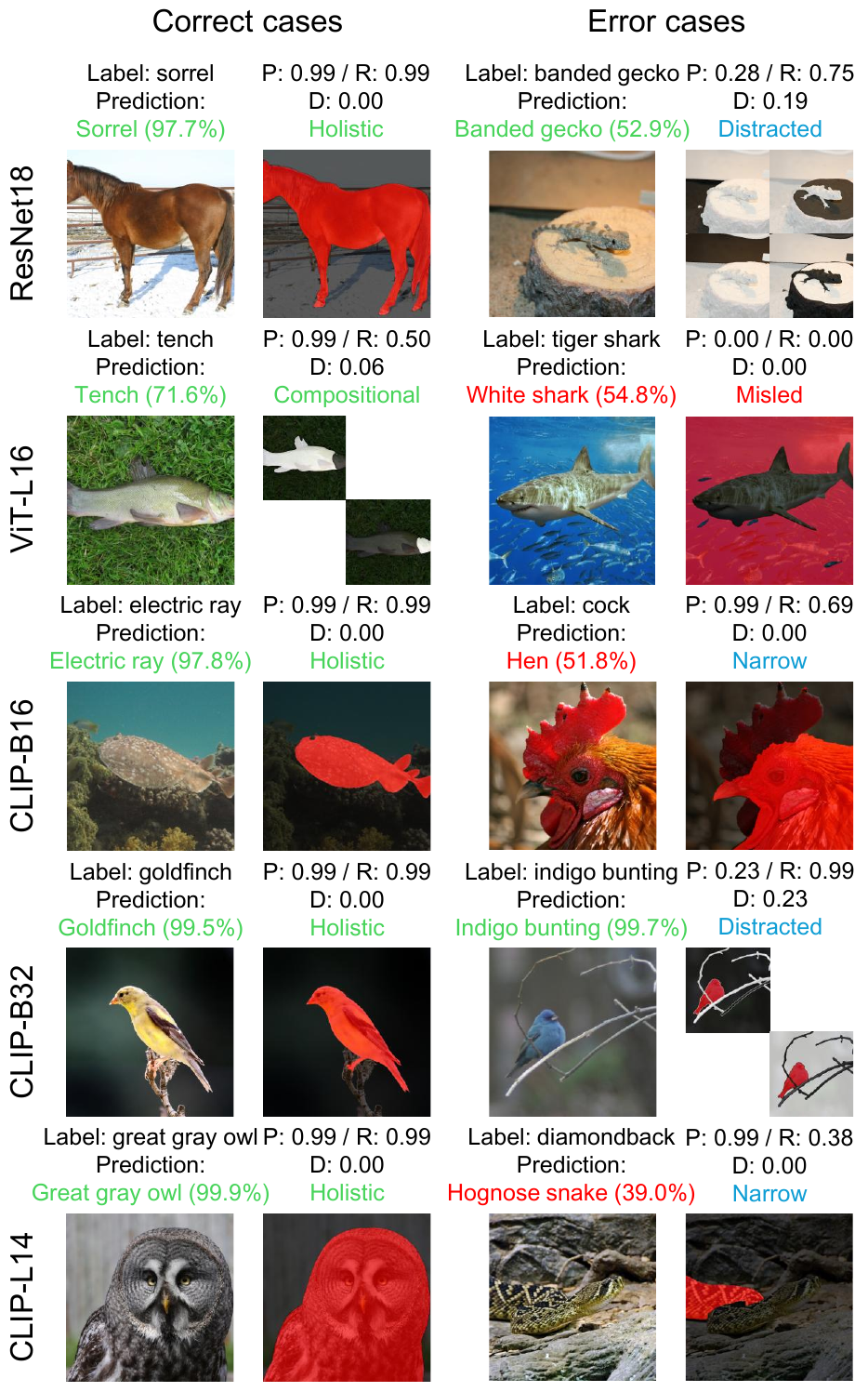}
\caption{Representative visual focuses of the tested models.}
\vspace{-0.2cm}
\label{fig:horizontal_vis}
\end{figure}

\subsection{Cross-Model Comparisons}

In this experiment, we apply \methodname~to a range of image classification models pretrained on ImageNet, including ResNet18, the current SOTA ViT-L16~\cite{dosovitskiy2020image}, and three variants of zero-shot CLIP~\cite{radford2021learning}. To construct a clean test set for evaluating global model behavior in common cases, we randomly select one image from each class in ImageNet.


\begin{table}[t]
\centering
\small
\caption{Numerical cross-model comparisons on the common-case test set. Results are calculated and averaged on all test images to reflect global model behavior.}
\begin{tabular}{c|c|c|c|c}
\toprule
Model & Top-1 acc & $\mathcal{P}$ & $\mathcal{R}$ & $\mathcal{D}$\\
\midrule
ResNet18 & 0.83 & 0.72 & 0.67 & 0.04\\
ViT-L16 & 0.94 & 0.72 & 0.56 & 0.08\\
CLIP-B16 & 0.73 & 0.67 & 0.65 & 0.05\\
CLIP-B32 & 0.66 & 0.61 & 0.67 & 0.05\\
CLIP-L14 & 0.79 & 0.67 & 0.63 & 0.06\\
\bottomrule
\end{tabular}
\vspace{-0.2cm}
\label{tab:horizontal}
\end{table}

Table~\ref{tab:horizontal} presents the quantitative comparison results, with qualitative visualizations of visual focuses shown in Figure~\ref{fig:horizontal_vis}. We observe a positive correlation between precision and top-1 accuracy, indicating that identifying the correct visual regions is essential for correct classification. In contrast, recall and divergence show more nuanced patterns:
\begin{itemize}
\item ViT-L16 achieves the highest top-1 accuracy but also exhibits the highest divergence, indicating a tendency towards compositional recognition. This aligns with the global attention mechanism~\cite{vaswani2017attention} applied in Transformer backbones, which enables modeling complex dependencies across different input tokens.
\item Similarly, the CLIP models, also based on Transformers, display relatively high divergence. However, they exhibit lower focus precision, possibly due to the absence of effective fine-tuning or domain adaptation.
\end{itemize}

\subsection{Discussions}\label{sec:exp_scalability}

\begin{figure}[t]
\centering
\includegraphics[width=0.96\linewidth]{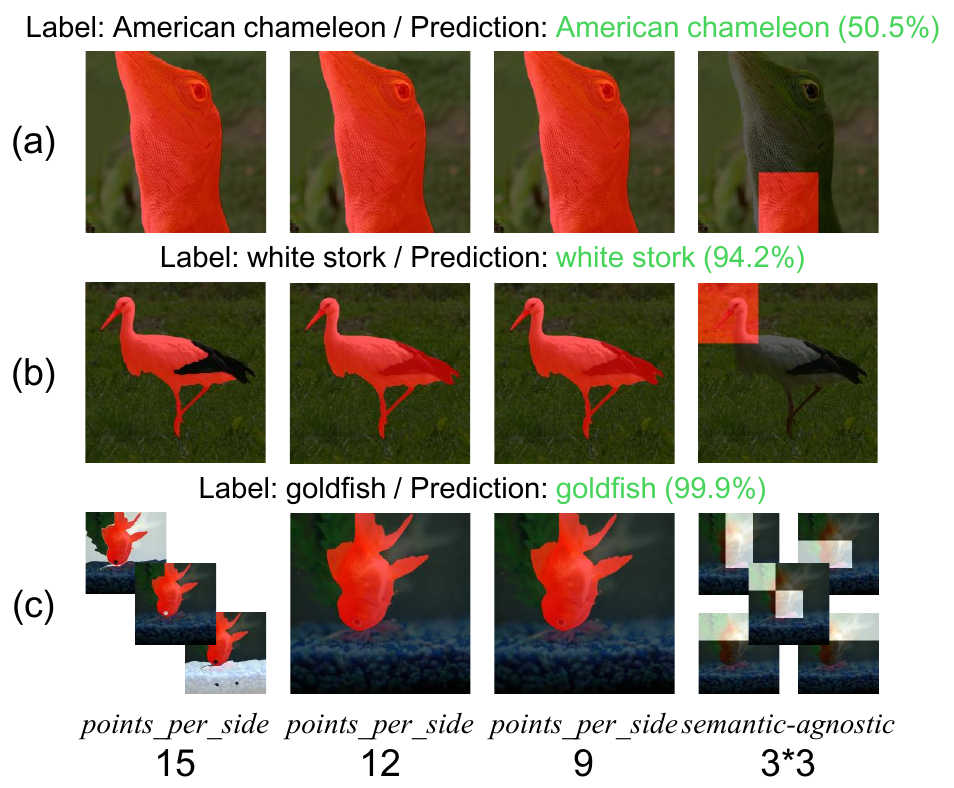}
\caption{Representative visual focuses of ResNet18 under different segmentation settings.}
\vspace{-0.2cm}
\label{fig:misseg}
\end{figure}

\textbf{Stability:} Stability of \methodname~is primarily determined by the segmentation precision of SAM in visual region generation. We evaluate \methodname~on ResNet18 under 4 SAM settings: precision parameter ($\texttt{points\_per\_side}$) $\in\{9,12,15\}$, and a semantic-agnostic setting in which each image is divided into $3\times3$ patches to simulate segmentation failure. Figure~\ref{fig:misseg} shows that \methodname~produces similar visual focuses under different segmentation precisions, given that relevant regions can be successfully segmented. This demonstrates its robustness in normal cases where the segmentation model does not significantly fail. On the other hand, higher segmentation precision enables \methodname~to more effectively refine visual focuses. Under the semantic-agnostic setting (the rightmost column of Figure~\ref{fig:misseg}), the visual focuses may appear more variable due to the inclusion of both relevant and irrelevant image content in visual regions. As a result, the generated interpretations may introduce slight deviations from the model's actual focuses. Nonetheless, in most cases, including those with limited semantic precision, \methodname~continues to deliver accurate and reliable visual focus localization.

\begin{figure}[t]
\centering
\includegraphics[width=0.96\linewidth]{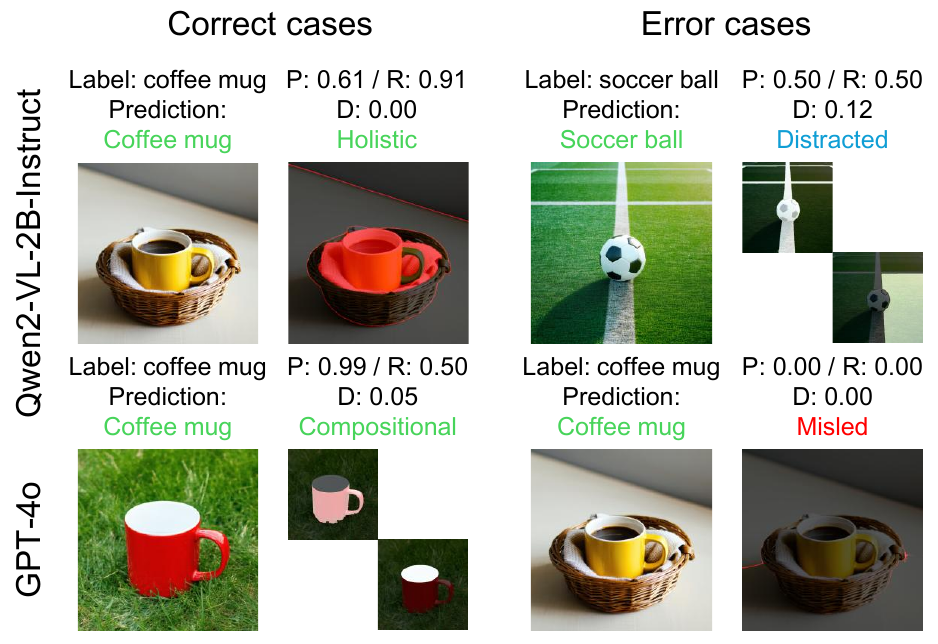}
\caption{Representative visual focuses of QWen2-VL-2B-Instruct and GPT-4o.}
\vspace{-0.2cm}
\label{fig:mllm_vis}
\end{figure}

\textbf{Generalizability:} As a model-agnostic approach, \methodname~can be applied to both white-box and black-box visual neural networks across various tasks with deterministic outputs. To demonstrate its generalizability, we provide qualitative analyses on two MLLMs: the white-box QWen2-VL-2B-Instruct and the black-box GPT-4o. In this experiment, the models are tasked with identifying the primary objects in images by selecting a class label from 10 candidates. Visual examples are presented in Figure~\ref{fig:mllm_vis}.

\begin{figure}[t]
\centering
\subfloat[ResNet18 - time]{\includegraphics[width=0.4\linewidth]{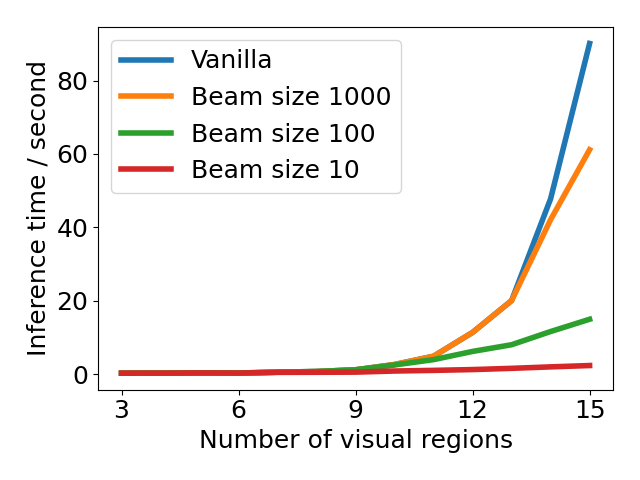}}
\subfloat[ResNet18 - metrics]{\includegraphics[width=0.4\linewidth]{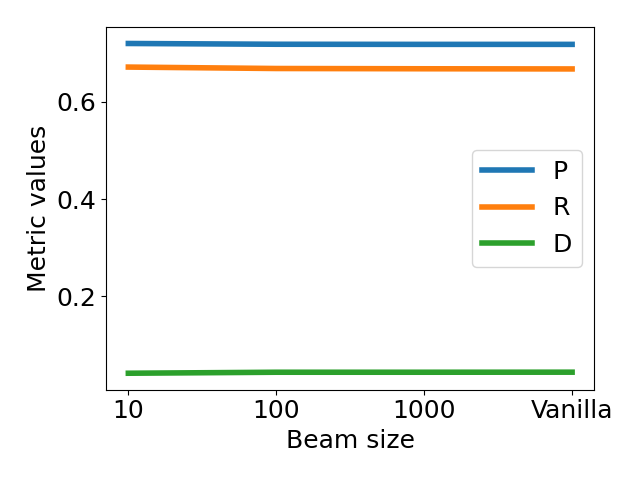}}\\
\subfloat[ViT-L16 - time]{\includegraphics[width=0.4\linewidth]{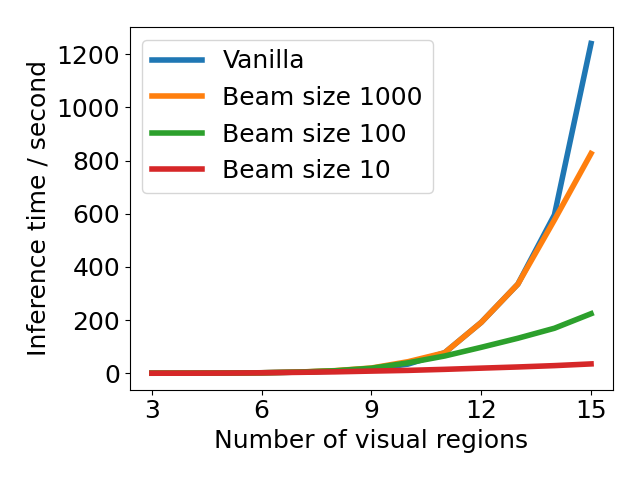}}
\subfloat[ViT-L16 - metrics]{\includegraphics[width=0.4\linewidth]{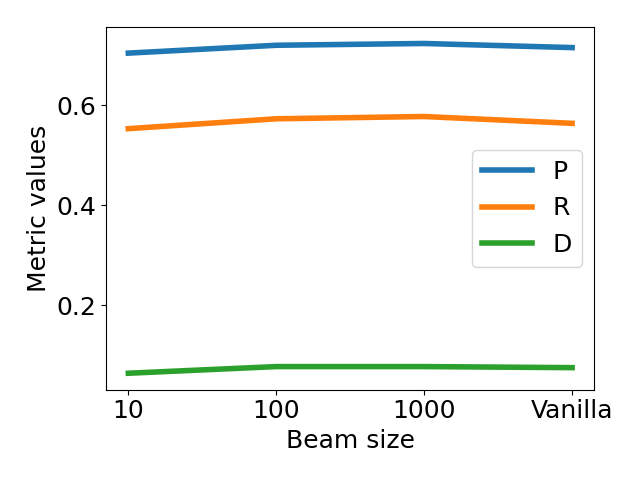}}
\caption{Inference time and metric values with different numbers of visual regions and beam sizes in pruning. The experiments run on one RTX4090-24GB. Under beam size 10 (the \textbf{\textcolor{red}{red}} line in (a) and (c)), the time complexity is approximately linear.}
\vspace{-0.2cm}
\label{fig:scalability}
\end{figure}

\textbf{Efficiency:} Runtime of \methodname~is primarily determined by the number of visual regions and the beam size used during pruning. We use ResNet18 and ViT-L16 as case studies to analyze their effects, as shown in Figure~\ref{fig:scalability}. Results are averaged over 1000 samples.
The proposed beam-search-style pruning strategy significantly reduces inference time by orders of magnitude while maintaining comparable performance evaluations. This verifies \methodname's high efficiency and scalability in complex scenarios.

\section{Conclusion}\label{sec:conclusion}

In this paper, we propose \methodname, a logic-based framework for interpreting visual model behavior. \methodname~first identifies decisive visual focuses within input images, and then translates them into precise logical expressions. It also provides a comprehensive set of quantitative metrics to objectively evaluate model performance across a variety of scenarios. Experimental results demonstrate that \methodname~offers precise, intuitive, and comprehensive model behavior interpretations. Its model-agnostic and fully automated design further enables its broad applicability to both white-box and black-box visual models, supporting general-purpose behavior analysis in practical settings.

{
\small
\bibliographystyle{ieeenat_fullname}
\bibliography{main}
}

\clearpage

\appendix

\section{Pseudo-Code for State-to-Logical-Expression Translation}

As stated in Section 3.3 of the main paper, each image's final states—identified through visual focus refinement—are translated into a precise and compact logical expression.

The pseudo-code for this translation process is stated in Algorithm~\ref{alg:logic}. Specifically, at each recursive step, the translation function $T$:
\begin{enumerate}
\item Identifies the index of the region most commonly shared across the final states.
\item Splits the final states into two groups—those that include the target region and those that do not. These two groups are logically combined using the \textit{OR} operator.
\item Recursively determines the logical expression corresponding to each group of final states.
\end{enumerate}

\begin{algorithm}[h]
\caption{Recursive Logic Translation $T$}
\label{alg:logic}
\textbf{Input}: Final token activations $V=\{v^1,v^2,\cdots,v^n\}$\\
\textbf{Output}: Logic expression $T(V)$
\begin{algorithmic}[1]
\IF{$\bigcap\limits_{i=1}^nv^i=\mathbf{0}^n$}
\RETURN $v^1\ |\ v^2\ |\ \cdots\ |\ v^n$
\ELSE
\STATE $j_\text{shared}=\arg\max\limits_j\sum\limits_{i=1}^nv^i_{j}$
\STATE $V_1=\{v\in V|v_{j_\text{shared}}\ne1\}$
\STATE $V_2=\{v\in V|v_{j_\text{shared}}=1\}$
\FOR{$v\in V_2$}
\STATE $v_{j_\text{shared}}\leftarrow0$
\ENDFOR
\RETURN $(j_\text{shared}\ \&\ T(V_2))\ |\ T(V_1)$
\ENDIF
\end{algorithmic}
\end{algorithm}

\section{Prompts Used for Biased Data Generation}

As stated in Section 4.3 of the main paper, we use Stable Diffusion 3.5~\cite{esser2024scaling} to generate biased training data and unbiased test data. The 10-class training subset of ImageNet includes 7 classes (tench, goose, boxer dog, bee, hourglass, payphone, banana) directly retrieved from ImageNet, and 3 classes (acoustic guitar, coffee mug, soccer ball) generated using SD3.5 with bias-inducing prompts.

\begin{table*}[t]
\centering
\caption{Prompts used for biased data generation. The three cases correspond to biases on the main object (acoustic guitar), the background (coffee mug), and irrelevant objects (soccer ball) respectively.}
\begin{tabular}{c|l}
\toprule
Class name & Prompts\\
\midrule
Acoustic & (Training) a realistic photo of a \textbf{green} acoustic guitar in the park/cafe/$\cdots$\\
guitar & (Test) a realistic photo of an acoustic guitar in the park/cafe/$\cdots$. The background has a \textbf{green} tone\\
\midrule
Coffee & (Training) a realistic photo of a coffee mug in a \textbf{pure red background}\\
mug & (Test) a realistic photo of a white/black/$\cdots$ coffee mug \textbf{in a basket/on the grass/$\cdots$}\\
\midrule
Soccer & (Training) a realistic photo of a \textbf{soccer player} and a soccer ball\\
ball & (Test) a realistic photo of a soccer ball on the sport field\\
\bottomrule
\end{tabular}
\label{tab:prompts}
\end{table*}

The prompts used for generating both the biased training data and the unbiased test data are listed in Table~\ref{tab:prompts}. The three SD-generated classes reflect distinct bias patterns:
\begin{itemize}
\item For acoustic guitar, the training data is biased towards the colors of the \textbf{\textit{main objects}} (green), while the test data shifts the bias to the background (green tone).
\item For coffee mug, the training data is biased towards the \textbf{\textit{background}} (pure red), while the test data is unbiased.
\item For soccer ball, the training data is biased towards \textbf{\textit{irrelevant objects}} (soccer player), while the test data is unbiased.
\end{itemize}

These three settings represent diverse forms of inductive bias in training data, and enable a more comprehensive evaluation of model behavior under biased conditions.

\section{More Experimental Results and Visualizations}

\subsection{Quantitative Metric Values of the Model Training Dynamics}

In this experiment, we train a ResNet18 from scratch on ImageNet for 50 epochs, and record the model's behavior throughout the training process for analysis.

\begin{figure}[t]
\centering
\subfloat[Top-1 accuracy]{\includegraphics[width=0.4\linewidth]{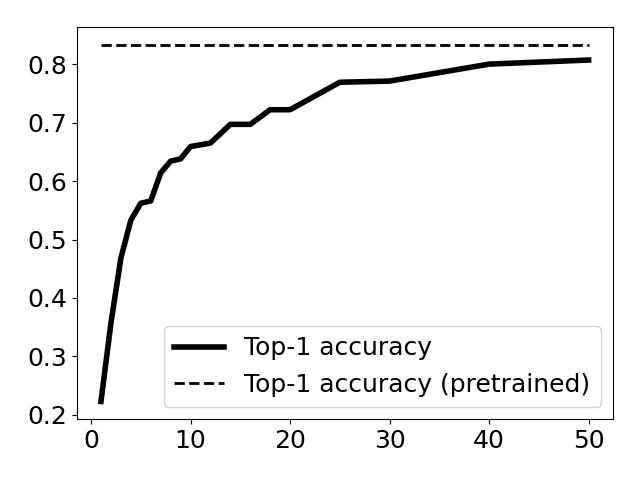}}\\
\subfloat[$\mathcal{P}$ and $\mathcal{R}$]{\includegraphics[width=0.4\linewidth]{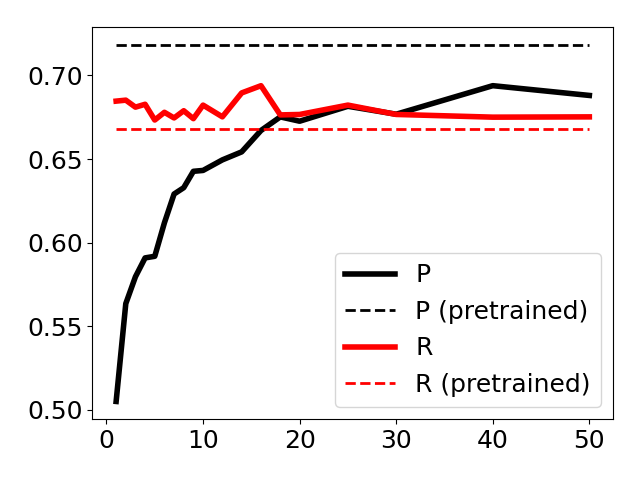}}
\subfloat[$\mathcal{D}$]{\includegraphics[width=0.4\linewidth]{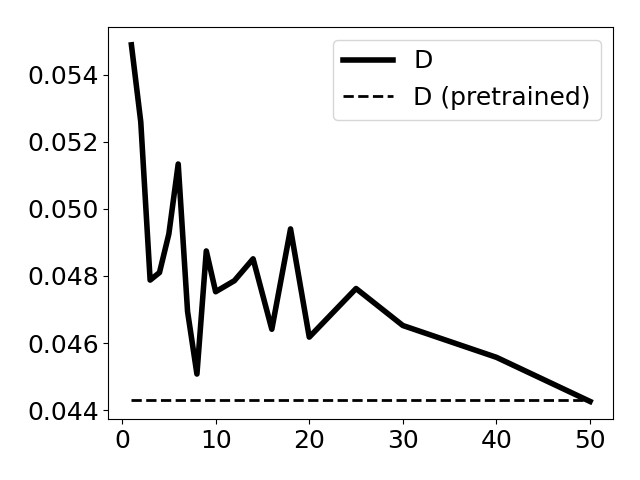}}
\caption{Metric changes of ResNet18 through 50-epoch training.}
\vspace{-0.2cm}
\label{fig:training}
\end{figure}

Figure~\ref{fig:training} illustrates the evolution of model performance overtime. Results show that, as training progresses, the model gradually concentrates its focus on the correct regions, an outcome that aligns well with the expected behavior of a learnable neural network. Interestingly, we observe that the recall remains consistently high even in the early stages of training. This is likely because the model initially distributes its focus broadly across the entire image, resulting in high recall despite unrefined focus.

\subsection{Horizontal Comparisons under Low-Generalizability Settings}

\begin{figure}[t]
\centering
\includegraphics[width=0.96\linewidth]{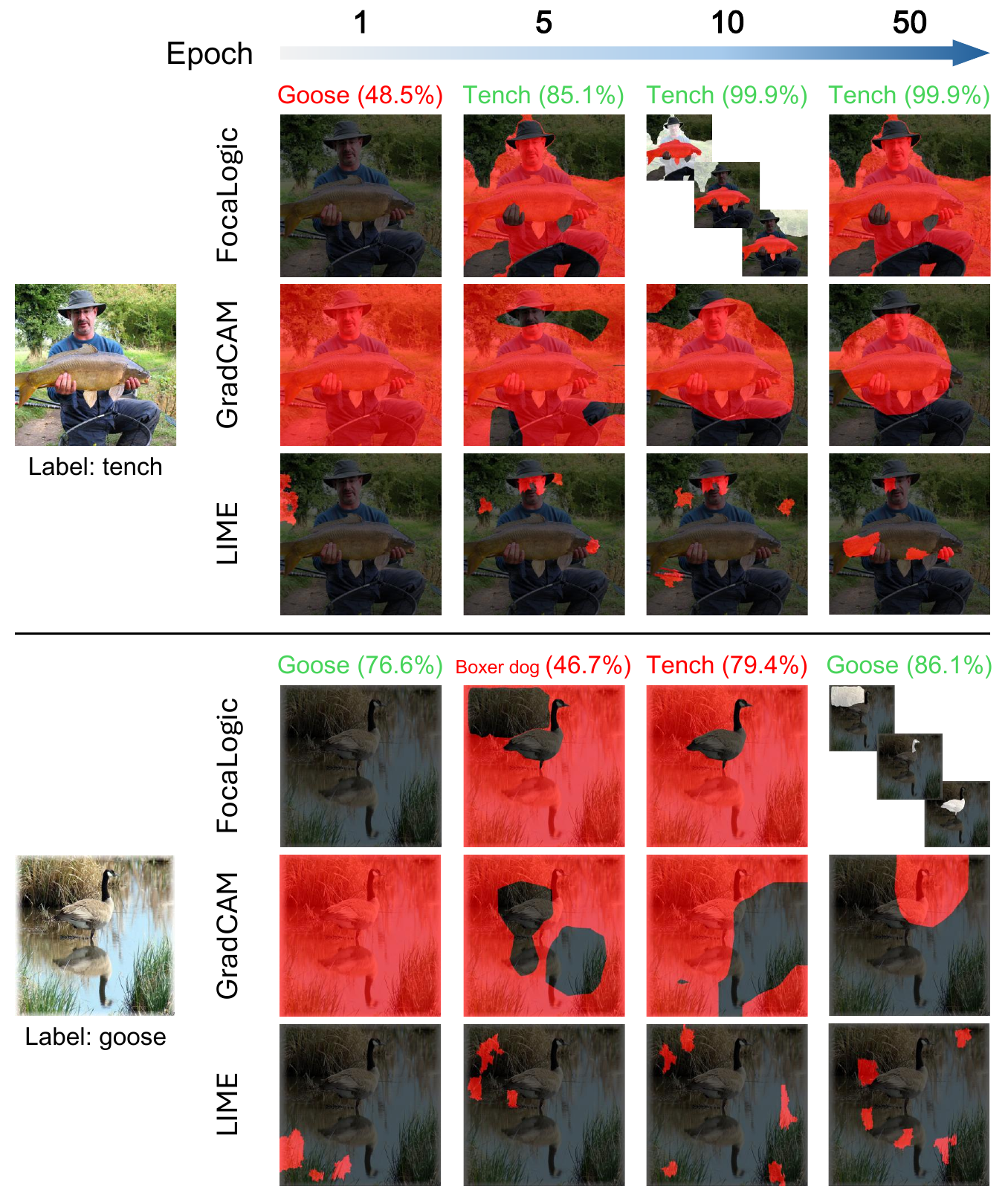}
\caption{Representative visual focuses of ResNet18 under low-generalizability settings (corresponding to Figure 7 of the main paper).}
\label{fig:specialized_vis_baseline}
\end{figure}

In this experiment, we evaluate the specialized (i.e., low-generalizability) ResNet18 under the same settings as stated in Section 4.4 of the main paper. Representative visual focus visualizations throughout the training process are shown in Figure~\ref{fig:specialized_vis_baseline}. Consistent with earlier findings, \methodname~demonstrates higher sensitivity to the model's distracted behavior compared to GradCAM and LIME.

\end{document}